%% file: main.tex
\documentclass[runningheads]{llncs}
\usepackage[table]{xcolor}

\usepackage{tikz} 
\usepackage{graphicx}
\usepackage{amsmath}
\usepackage{amssymb}
\usepackage{booktabs}
\usepackage{physics}
\usepackage{pifont}
\usepackage{colortbl}
\usepackage{multirow}
\usepackage{subcaption}
\usepackage{hhline}
\usepackage{comment}
\usepackage{tablefootnote}

\usepackage[accsupp]{axessibility}  

\usepackage{subfiles}
\input{00_includes}
\usepackage[pagebackref,breaklinks,colorlinks]{hyperref}

\begin{document}
\pagestyle{headings}
\mainmatter
\def\ECCVSubNumber{5669}  

\title{\LONGOURS}

\titlerunning{\OURS{}}

\author{
M. J. Tyszkiewicz\inst{2}\thanks{Work done while at Google Research.} \and
K.-K. Maninis\inst{1} \and
S. Popov\inst{1} \and
V. Ferrari\inst{1}
}
\authorrunning{Tyszkiewicz et al.}
\institute{$^1$Google Research \qquad $^2$EPFL}
\maketitle

\begin{abstract}
We propose a transformer-based neural network
architecture for multi-object 3D reconstruction from RGB
videos. It relies on two alternative ways to represent its knowledge: as a
global 3D grid of features and an array of view-specific 2D grids. We
progressively exchange information between the two with a dedicated
bidirectional attention mechanism. We exploit knowledge about the image
formation process to significantly sparsify the attention weight matrix,
making our architecture feasible on current hardware, both in terms of memory and
computation.
We attach a DETR-style head~\cite{carion2020end} on top of the 3D
feature grid in order to detect the objects in the scene and to predict their 3D pose and 3D shape.
Compared to previous methods, our architecture is
single stage, end-to-end trainable, and it can reason holistically about a
scene from multiple video frames without needing a brittle tracking
step.
We evaluate our method on the challenging Scan2CAD dataset~\cite{avetisyan19cvpr},
where we outperform
(1) state-of-the-art methods~\cite{maninis20vid2cad,li21odam,li2020arxiv,rukhovich2021imvoxelnet} for 3D object pose estimation from RGB videos;
and (2) a strong alternative method combining Multi-View Stereo \cite{duzceker21deepvideomvs} with RGB-D CAD alignment \cite{avetisyan19iccv}.
\end{abstract}

\input{01_introduction}

\input{02_related}
\input{03_method}
\input{04_experiments}

\input{05_conclusions}

\input{06_appendix}
\clearpage
{\small
\bibliographystyle{splncs04}
\bibliography{references}
}
\clearpage

\end{document}

%% file: 00_includes.tex
\newcommand{\OURS}{RayTran}

\newcommand{\LONGOURS}{\OURS: 3D pose estimation and shape reconstruction of multiple objects from videos with ray-traced transformers}

\newcommand{\rott}[1]{\rotatebox[origin=lB]{90}{#1}}

\newcommand{\mypar}[1]{\hspace{-6.5mm} \textbf{#1.}}
\newcommand{\myfirstpar}[1]{\textbf{#1.}}

\newcommand{\stimes}{{\mkern-2mu\times\mkern-2mu}} 

%% file: 01_introduction.tex

\section{Introduction}
\label{sec:intro}

\begin{figure*}
	\centering
	\vspace{-2mm}
	\includegraphics[width=1\linewidth]{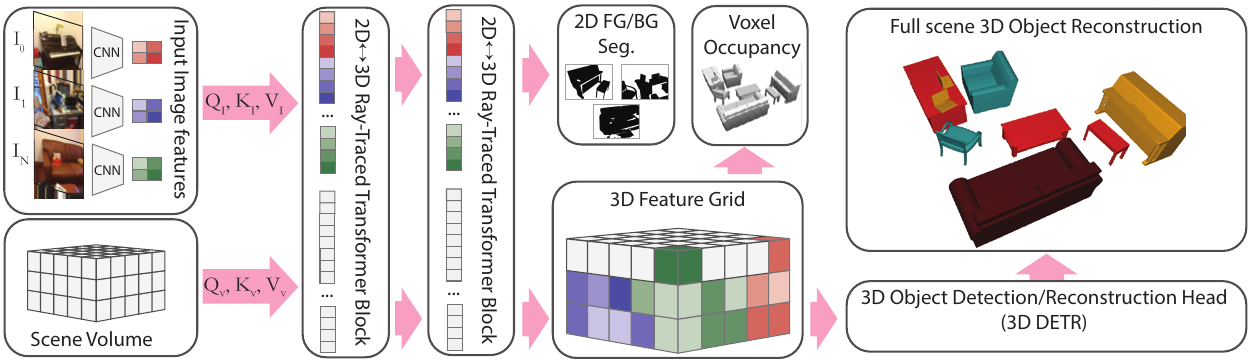} \\[1mm]
	\vspace{-2mm}
	\caption{\small
	\textbf{Overview of our method:}
	The \OURS{} backbone processes information in two parallel network streams. The first one (2D) works on features extracted on the multiple input frames. The second one (3D) starts from an empty volumetric feature representation of the scene. The 2D stream gradually consolidates features on the 3D volume and visa-versa with repeated blocks of ray-traced transformers.
	The backbone outputs a 3D feature grid which offers a global representation of the 3D volume of the scene.
	We attach a DETR-style head~\cite{carion2020end} to this representation, to detect all objects in 3D and
	to predict their 3D pose and 3D shape.
	We further help training with two auxiliary tasks:
	predicting 3D coarse binary occupancy for all objects together, and predicting amodal 2D foreground-background masks.
	}
	\label{fig:overview}
	\vspace{-4mm}
\end{figure*}

Detecting and reconstructing objects in 3D is a challenging task with multiple applications in computer vision, robotics, and AR/VR that require semantic 3D understanding of the world.
%
In this paper we propose~\OURS{}, a transformer-based~\cite{vaswani2017attention}
neural network architecture for reconstructing multiple objects in 3D given an RGB video as input.
Our key new element is a backbone which infers a global representation of the 3D volume of the scene.
We attach a DETR-style head~\cite{carion2020end} on top of it, which detects objects in the 3D representation and predicts their 3D pose and shape (Figure~\ref{fig:overview}).

The backbone inputs multiple video frames showing different views of the same static scene.
Its task is to jointly analyze all views and to consolidate the extracted
information into a global 3D representation.
Internally, the backbone maintains two alternative scene representations.
The first is three-dimensional and describes the volume of the scene.
The second is two-dimensional and describes the volume from the perspective of the individual views.
We connect these two representations with a bidirectional attention mechanism to exchange information between them, allowing the 3D representation to progressively accumulate view-specific features, while at the same time the 2D representation accumulates
global 3D features.

Processing videos with transformers is notoriously
resource-consuming~\cite{bertasius2021timesformer,arnab2021vivit}.
Our case is no exception: if we relied on attention between
all elements in the 2D and 3D representations,
the attention matrix would have infeasible memory requirements
(and it would also be computationally very expensive).
To overcome this, we propose a \emph{sparse ray-traced attention} mechanism.
Given the camera parameters for each view, we exploit the image formation process to identify pairs of
2D and 3D elements that are unlikely to interact.
We omit these pairs and store the attention matrix in a sparse format.
This greatly reduces its computational and memory complexity, by a factor of
$O(|V|^\frac{2}{3})$, where $|V|$ is the number of voxels in the 3D representation.

We attach a DETR-style head~\cite{carion2020end} on top of the 3D representation produced by the backbone. This head detects objects and predicts their class, 3D shape, and 3D pose (translation, rotation, scale).
We represent object shapes with a voxel grid and then extract meshes using marching cubes~\cite{lewiner03jgtools}.
We also predict coarse binary volumetric occupancy for all objects together, using a 3D convolutional layer on top of the global 3D representation.
This provides an auxiliary task that teaches the network about the scene's geometry, and is essential for  training.

As a second auxiliary task, we add an additional network head that predicts the 2D amodal foreground-background binary masks of all objects in the scene.
Besides enabling this task, this head also helps training the backbone as it closes the loop between images and the 3D representation.

Several recent works~\cite{maninis20vid2cad,runz20cvpr,li21odam} tackle
3D scene reconstruction from videos in the same setting. They 
rely on a 3-step pipeline:
(1) object detection in individual 2D frames, along with estimating properties such as 3D rotation, parts of 3D scale, and 3D shape (either as a parametric surface~\cite{li21odam} or by retrieving a CAD model from a database~\cite{maninis20vid2cad});
(2) tracking-by-detection~\cite{andriluka2008people,breitenstein2009robust,bergmann2019tracking}, to associate 2D detections across frames;
(3) multi-view optimization to integrate the per-frame predictions. This completes
all 3D pose parameters, resolving the scale-depth ambiguities, and places all objects in a common, global 3D coordinate frame.

Our method was inspired by these works and  addresses several of their
shortcomings. The pipelines are composed of heterogeneous steps,
which are trained separately and require manual tuning to work well together.
The pipelines are complicated and over-engineered due to the intricate nature of the full-scene
object reconstruction task.
The tracking step is especially brittle. Objects often go out of view and re-appear later, and occlude each other over time. This poses a major challenge and leads to objects broken into multiple tracks, as well as tracks mixing multiple objects. These tracking errors harm the quality of the final 3D reconstructions.

In contrast, our method is end-to-end trainable. It is built from well
understood neural network modules and it has a simple, modular architecture in comparison.
Importantly, {\em we avoid tracking altogether}.
Furthermore, our method does not rely on any notion of time sequence, so it is also applicable to sparse multi-view inputs (in addition to video).

%
We evaluate \OURS{} on the challenging Scan2CAD~\cite{avetisyan19cvpr} dataset, featuring videos of complex indoor scenes with multiple objects.
Through extensive comparisons we show that \OURS{} outperforms several works:
(1) two baselines that process frames individually, defined in~\cite{maninis20vid2cad} as extensions of Mask2CAD~\cite{kuo20eccv}. This illustrates the value of jointly processing multiple frames in \OURS{};
(2) four recent multi-frame methods Vid2CAD~\cite{maninis20vid2cad}, ODAM~\cite{li21odam}, MOLTR~\cite{li2020arxiv}, ImVoxNet~\cite{rukhovich2021imvoxelnet}.
Besides performing better, \OURS{} also offers a much simpler design than~\cite{maninis20vid2cad,li21odam,li2020arxiv}, with an end-to-end trainable, unified architecture which does not require a tracking module;
(3) a strong alternative method that combines the state-of-the-art Multi-View Stereo~\cite{duzceker21deepvideomvs} and RGB-D CAD alignment~\cite{avetisyan19iccv} methods.

%% file: 02_related.tex
\section{Related Work}
\label{sec:related}

\myfirstpar{3D from multiple views}
Classic SfM/SLAM works cast 3D reconstruction as estimation of 3D points from multiple views based on keypoint correspondences~\cite{pollefeys99ijcv,mur15transrobotics,wu133dv,schonberger16cvpr,engel2017direct}.
However, the output point cloud is not organized into objects instances with their classes, 3D shapes, or poses.
A line of works detect and localize objects in 3D using multi-view projection constraints, by approximating the object shapes with 3D boxes~\cite{yang2019cubeslam} and ellipsoids~\cite{nicholson2018quadricslam}. ODAM~\cite{li21odam} goes a step further to creates a scene representation out of superquadrics, by using a graph neural network as core architecture for object association in time.
FroDO~\cite{runz20cvpr} and MO-LTR~\cite{li2020arxiv} rely on both 2D image cues and the sparse 3D point clouds from SfM/SLAM to reconstruct objects in the scene.
Qian et al.~\cite{qian20eccv} produce volumetric reconstructions of multiple objects in a 
synthetically generated scene.
Vid2CAD~\cite{maninis20vid2cad} integrates the single-view predictions of Mask2CAD~\cite{kuo20eccv} across time, to place objects from a CAD database into the 3D scene.

A common caveat of multi-view methods for 3D object reconstruction is that their
architectures are overly complex, they cannot be trained end-to-end due to their heterogeneity, and they often rely on a brittle tracking-by-detection step. Instead, our proposed method provides a light-weight end-to-end architecture for the task, while we completely avoid tracking.

Similar to \OURS{}, the concurrent ImVoxelNet~\cite{rukhovich2021imvoxelnet} keeps its 3D knowledge in a global 3D representation and does not require tracking.
It uses a hand-crafted unidirectional mechanism to project and consolidate image features onto it.
In contrast, our ray-traced transformers learn the optimal way to consolidate features. They are also bidirectional, which enables 2D supervision through re-projection as well as additional tasks, like novel-view synthesis.
Moreover, \OURS{} reconstructs the 3D shapes of the detected objects, going beyond detecting 3D boxes.

\mypar{Transformer architectures for computer vision}
Several recent works use attention-based architectures (transformers)~\cite{vaswani2017attention} for computer vision tasks.
ViT~\cite{dosovitskiy2020vit} replaces the traditional convolutional backbones with attention among patches for image classification. The same idea has been incorporated into network designs for semantic segmentation~\cite{strudel2021segmenter,zheng2021rethinking,cheng2021maskformer}, object detection~\cite{carion2020end}, and panoptic segmentation~\cite{cheng2021maskformer}.
Transformers have been introduced recently also for video processing. 
TrackFormer~\cite{meinhardt2021trackformer} uses a transformer architecture for multi-object tracking. ViViT~\cite{arnab2021vivit}  and TimeSFormer~\cite{bertasius2021timesformer} use ViT-like patches from multiple frames for video classification.

The main bottleneck of these approaches are the prohibitive memory requirements. TrackFormer~\cite{meinhardt2021trackformer} can only process 2 images at a time, which prevents end-to-end training on the whole video. Similarly, the all-to-all patch attention, which is the cornerstone of~\cite{bertasius2021timesformer,arnab2021vivit}, comes with often infeasible memory requirements. ViViT~\cite{arnab2021vivit} needs the combined memory of 32 TPU accelerators to process a single batch of 128 frames.
Our work overcomes these limitations by using sparse attention between 2D and 3D features. The sparsity is achieved by using image formation constraints directly from the poses of the cameras, which significantly reduces the memory requirements. For reference, \OURS{} processes up to 96 frames of a video and reconstructs all instances on a single 16 GB GPU. 

\mypar{3D using a dedicated depth sensor}
Our work draws inspiration from several 3D object reconstruction methods that directly work on point clouds obtained by fusing RGB-D video frames.
Early works use known pre-scanned objects~\cite{salas13cvpr}, hand-crafted features~\cite{nan2012tog,frome04eccv,li2015database,shao2012tog}, and human intervention~\cite{shao2012tog}.
Recent works use deep networks to directly align shapes on the dense point clouds~\cite{avetisyan19cvpr,avetisyan19iccv,avetisyan20eccv,izadinia20cvpr,shan2021ellipsdf}.
Fei et al.~\cite{fei18eccv} align a known set of shapes on a video in 4 DoF, by using a camera with an inertial sensor.

Using an additional sensor reduces the search-space required to accurately re-construct an object in 3D. Both the depth and the inertial sensors eliminate the depth-scale ambiguity, and compared to re-constructing from pure RGB, RGB-D sensors provide cleaner, much more realistic results.
Our work does not require the intermediate step of point-based reconstruction,
does not use the extra depth sensor, and can directly reconstruct objects in a posed RGB video.

\mypar{3D detection and reconstruction from a single image}
Pioneering works in this area process a single image to either infer the pose of an object as an oriented 3D bounding box~\cite{mousavian20173d,mahendran2018mixed}, or to also predict the 3D shape of the object~\cite{wang18eccv,mescheder19cvpr,choy16eccv,girdhar16eccv,wu16nips,xie20ijcv,park19cvpr,chen20cvpr}.
Works that are able to predict an output for multiple object instances, typically first detect them in the 2D image, and then reconstruct their 3D pose and/or shape~\cite{huang18eccv,gkioxari19iccv,kuo20eccv,izadinia17cvpr,tulsiani18cvpr,kundu18cvpr,nie20cvpr,popov20eccv,kuo2021patch2cad,engelmann2021points,hu2019joint,gumeli2021roca}.

3D predictions from single images tend to be inaccurate due to scale-depth ambiguity, and often methods of this category compensate for it in a variety of ways, e.g., based on estimating an approximate pixel-wise depth map from the input image~\cite{huang18eccv}, by requiring manually provided objects' depth and/or scale~\cite{gkioxari19iccv,kuo20eccv,kuo2021patch2cad} at test time,
or by estimating the position of a planar floor in the scene and assuming that all objects rest on it~\cite{izadinia17cvpr}.
Some works~\cite{tulsiani18cvpr,nie20cvpr,popov20eccv} attempt to predict object depth and scale directly based on image appearance.
Our proposed approach processes multiple frames simultaneously, and implicitly compensates for the scale-depth ambiguity by using many different view-points of the objects appearing in the scene.

%% file: 03_method.tex

\vspace{-2mm}
\section{Proposed Approach}
\label{sec:method}
\vspace{-2mm}

Our method takes multiple views (video frames) of a scene and their camera parameters as input.
Each view captures a different part of the same 3D scene. It outputs
the 3D pose (rotation, translation, scale), the class, and the 3D shape of all objects in the scene.

We achieve this with a single, end-to-end trainable, neural network model. 
We propose a transformer-based backbone that processes the input views and infers a
global 3D volume representation for the entire scene. We use this representation to predict the object shapes, poses, and classes, by attaching a DETR-style~\cite{carion2020end} head to it.
In addition, we perform two additional auxiliary tasks: 3D occupancy, where we predict coarse binary volumetric occupancy for all objects together, and 2D foreground-background amodal segmentation.
The overview of our architecture is illustrated in Fig.~\ref{fig:overview}.

\vspace{-2mm}
\subsection{The \OURS{} Backbone}

\begin{figure}
	\centering
	\vspace{-2mm}
	\includegraphics[width=.44\linewidth]{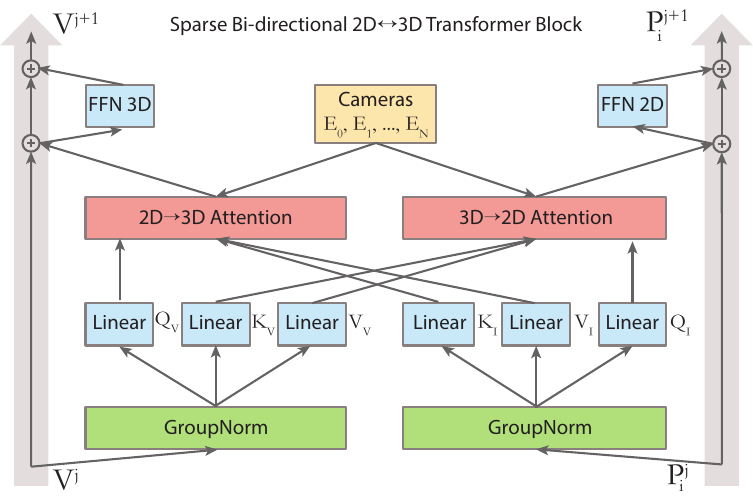}
	\hspace{1em}
	\includegraphics[width=.49\linewidth]{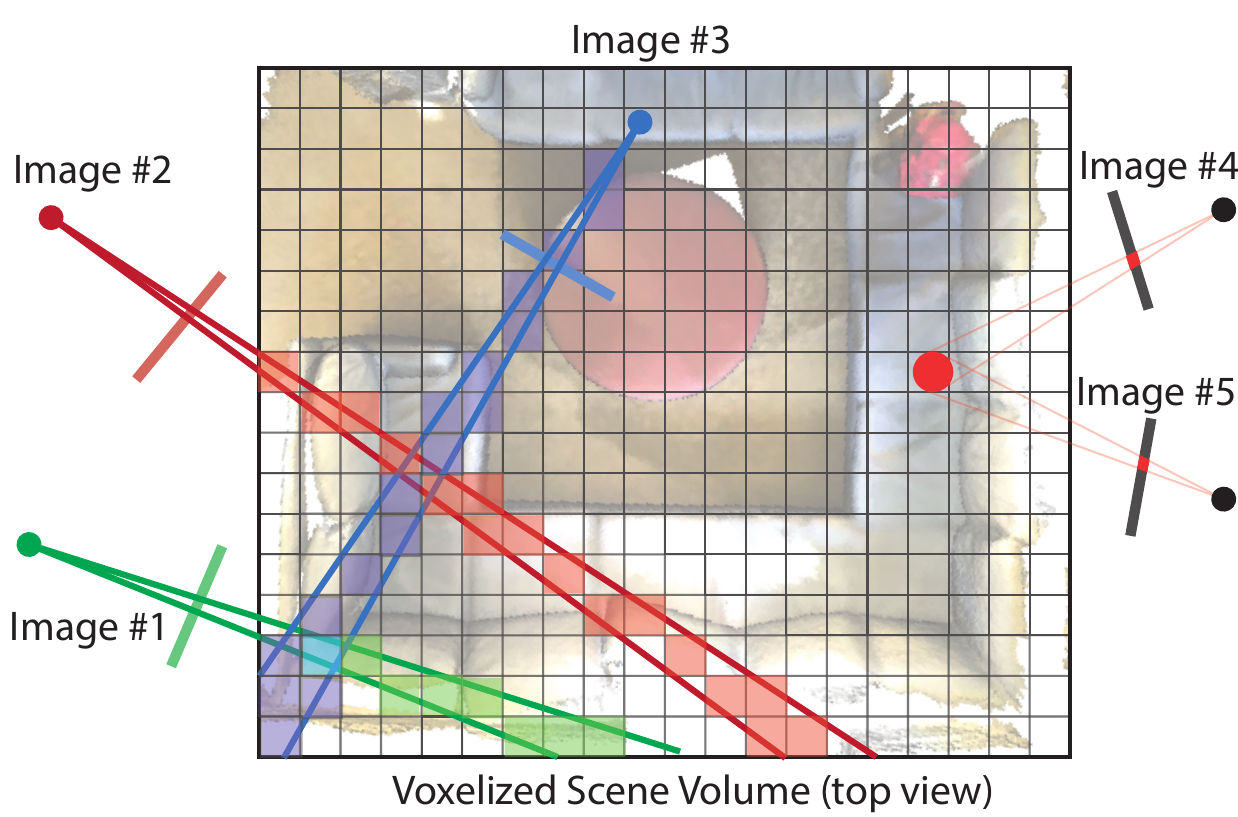} \\[1mm]
	\vspace{-2mm}
  \caption{\small
    $\boldsymbol{2D \Leftrightarrow 3D}$ \textbf{ray-traced transformer block (left).}
    Each block uses two parallel residual network streams that exchange information by attention.
    They consist of two layers of ray-traced sparse attention (2D$\rightarrow$3D and 3D$\rightarrow$2D) followed by a feed-forward network (FFN) composed of 3D and 2D convolutions, respectively. The voxel features (3D) inform the image features (2D) at each stage of the backbone.
    The 3D reconstruction head uses the voxel features (output of left stream), whereas the 2D foreground-background segmentation head uses the pixel grid (output of right stream). \\
    \textbf{Intertwining 3D voxel- with 2D image features (right):} Multiple voxels can project on the same pixel, and multiple pixels from multiple cameras can look at the same voxel. The proposed attention layer models this interaction in an intuitive way.
  }
	\label{fig:attention_layer}
	\vspace{-4mm}
\end{figure}

We propose a neural network architecture that operates on two alternative
representations in parallel. The first one is three-dimensional and describes the 3D space
that the scene occupies. We use a voxel grid $V$ with \emph{global} features that
coincides with this space. The second one is two-dimensional and describes the
scene from the perspective of the individual views. For each view $i=1..N$, we use
a pixel grid $P_i$ of \emph{image} features that coincides with the view's
image.

The two representations are connected implicitly through the image formation
process. We model this as a sequence of $2D \Leftrightarrow 3D$ neural network
transformer blocks (Figure~\ref{fig:attention_layer}, left). The $j$-th block
takes all views $P^j_{1...N}$ and the volume $V^j$ as input, mixes their features, and outputs a pair of new
representations ($P^{j+1}_{1...N}$ and $V^{j+1}$).
This allows the global 3D representations to be progressively populated by local
features from the different views, while at same time the 2D representations
progressively accumulate global features in different depths of the network.

The output of the \OURS{} backbone is a 3D feature representation of the scene, derived from the input views.
In order to compute the initial 2D representation $P^0_i$, we embed ResNet-18~\cite{he16cvpr} in our backbone (pre-trained on ImageNet). We run ResNet-18 over the input views $i$ and we take the output of its last block for each view. 
To initialize the 3D volume representation $V^0$, we cast a ray (un-project) from all the pixels $P^0_i$ onto the 3D volume.
We then average the image features that fall into each voxel of $V^0$.

\mypar{Block operation}
The 2D$\Leftrightarrow$3D blocks of \OURS{} consist of two parallel network streams, as shown in
Figure~\ref{fig:attention_layer} (left). The first one ($2D \Rightarrow 3D$),  mixes
features from $P^{j}_i$ into $V^{j}$ and outputs $V^{j+1}$. The second one ($3D
\Rightarrow 2D$) from $V^{j}$ into $P^{j}_i$, resulting in $P^{j+1}_i$. We
propose to build both networks using the multi-headed attention
mechanism~\cite{vaswani2017attention}.

The attention mechanism can translate an input vector (1D array of features) from a source domain into a differently-sized vector in a target domain. To do this, the mechanism computes a \emph{key} vector that describes each position in the source domain and a \emph{query} vector that describes each position in the target domain. It then computes a matrix describing the relation between source and target positions, by storing the dot product between the features at position $i$ in the \emph{key} and position $j$ in the \emph{query} at $(i, j)$ in the matrix.
Finally, the mechanism computes a \emph{value} vector from the input vector and multiplies this with the attention matrix in order to obtain the output.
The \emph{key} and the \emph{value} depend on the input vector (from the source domain), while the \emph{query} 
depends on a vector from the target domain. The goal of the attention mechanism is to learn the dependencies between the two domains.

The attention mechanism is intrinsically well suited to model the connection between pixels
and voxels. Multiple pixels from multiple cameras can look at the same voxel, as
shown in Fig.~\ref{fig:attention_layer} (right). We need a mechanism to consolidate their
features in the voxel. Similarly, multiple voxels can project onto the
same pixel and we need to consolidate their features. The matrix-\emph{value}
multiplication in the attention mechanism naturally achieves the desired effect. 

For $2D \Rightarrow 3D$ attention, we derive the \emph{key} and the \emph{value} from all pixels
from all views of $P^j_i$ and the \emph{query} from all voxels of $V^j$. For $3D \Rightarrow 2D$
attention, conversely, from $V^{j}$ and $P^j_i$.
We introduce skip connections in both networks, by adding the inputs of the
attention mechanism to its outputs. We then post-process with a feed-forward network, built with 3D and 2D convolution layers respectively (Fig.~\ref{fig:attention_layer}).

\mypar{Ray-traced attention layers}
In a realistic setting, the attention layer has infeasible memory requirements.
Our backbone operates on multiple frames simultaneously,
20 during training and 96 at inference time, each using a 2D feature grid of $40 \times 30$ for the 2D features $P_i$. 
We use a voxel grid with resolution $48 \times 48 \times 16$ to model a $9m \times 9m \times 3.5m$ volume, 
corresponding to voxel dimensions of approximately $19cm \times 19cm \times 22cm$. 
We use 256 features in both the 2d and 3d representations and 8 heads in the attention layers.
Given the above numbers, the attention matrices in each 2D $\Leftrightarrow$ 3D block alone would require
$\approx 52\textrm{GB}$ of memory with 20 frames, which is prohibitive.

\begin{figure}[t]
	\centering
	\vspace{-2mm}
	\includegraphics[width=1\linewidth]{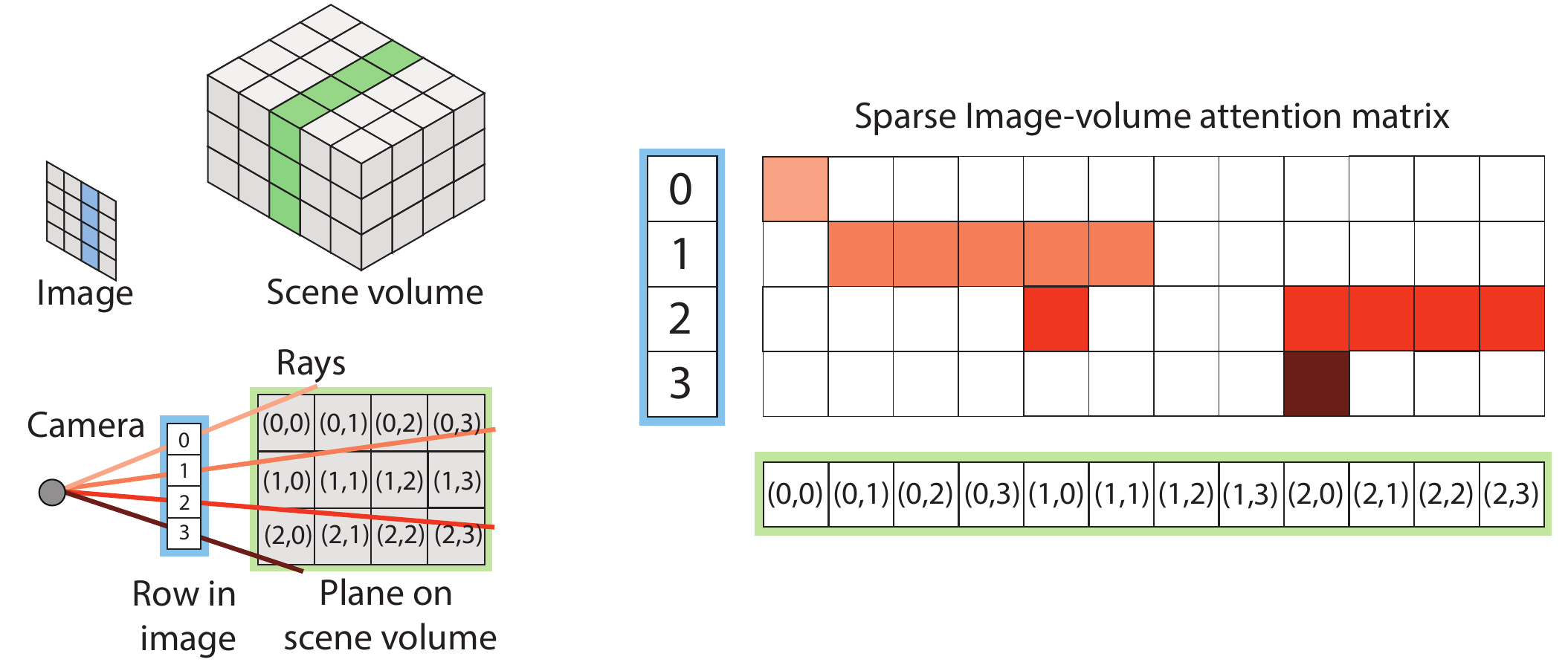} \\[1mm]
	\vspace{-4mm}
	\caption{\small
	\textbf{Ray-traced sparse attention.}
    A pixel and a voxel are likely to interact only if a ray passes through both of them.
    We exploit this to significantly reduce the memory requirements of our 2D $\Leftrightarrow$ 3D blocks, by
    sparsifying the attention matrices.
    If no ray passes though a pixel/voxel pair, the two are unlikely to interact and we omit the corresponding value in the matrix.}
	\label{fig:sparse_attention}
	\vspace{-4mm}
\end{figure}

To overcome this, we embed knowledge about the image
formation process into the architecture (Fig.~\ref{fig:sparse_attention}).
A pixel and a voxel can interact with each other directly only if there is a camera ray that passes
through both of them. If no such ray exists, the two are unlikely to interact, and we set
the corresponding entry in the attention matrix to zero. This is mathematically
equivalent to the masking mechanism employed in autoregressive transformers to enforce
causality~\cite{vaswani2017attention}, but crucially allows us to store the matrix
in sparse form and significantly reduce memory consumption.
A pixel can only interact with $O(\sqrt[3]{|V|})$ voxels, where $|V|$ is the number of voxels in
$V$, since any ray can only pass through at most this many voxels.
We thus need $O(|V|^\frac{2}{3})$ times less memory to store the matrix. In
sparse coordinate format, which encodes each matrix entry with 3 numbers (row, column, value), the matrix from our
example above would consume $270$ times less memory ($3\times64.4\textrm{MB}$
instead of $52\textrm{GB}$).
We call multi-headed attention based on such sparse matrices \emph{ray-traced sparse attention}.

We use the camera parameters to determine which pixel-voxel pairs interact with each other. In turn, the camera parameters can be computed with off-the-shelf pipelines such as COLMAP~\cite{schonberger16cvpr}.
To make full use of the
limited volume that our backbone can focus on, we center the camera positions
within it.

\vspace{-2mm}
\subsection{Task-specific heads on top of the backbone}
\label{sec:heads}

\myfirstpar{3D pose estimation and shape reconstruction}
For our main task, we predict the 3D pose, and reconstruct the shape of all objects
seen in the video. We use a
DETR-style~\cite{carion2020end} architecture, with multi-headed attention between
64 object query slots and the voxels in the backbone output. In each slot, we
predict the object's class, its shape in canonical pose, 3D center, 3D anisotropic scale,
and 3D rotation. We use a special \emph{padding} class to indicate that a query
slot does not contain a valid object. We encode the shape as a
$63\stimes63\stimes63$ voxel grid, which we predict with a sequence of
transposed 3D convolution layers from the query's embedding. We use Marching
Cubes~\cite{lewiner03jgtools} to convert the voxel grid to a mesh.
We only predict rotation around the `up'-axis of each object (as one angle), as most objects in our dataset are only rotated along this axis.

We use cross entropy for predicting the class, binary cross entropy for the shape's
voxels, soft $L_1$ loss for the object center, $L_1$ loss over the logarithm of the
scales, and a soft $L_1$ loss for the rotation angles. Finally, we match predictions to
ground-truth objects in DETR using a linear combination of all losses except the voxel one,
which we exclude for performance reasons. As in~\cite{carion2020end}, we supervise at all intermediate layers.

\mypar{3D occupancy prediction}
As an auxiliary task, we also predict the binary 3D occupancy of all objects for the whole scene on a coarse voxel grid. 
We use one 3D convolution layer on
top of the backbone output, and we supervise with binary-cross-entropy.
We run occupancy prediction as an auxiliary task, to directly teach the network about the combined object geometry.
This is crucial for the 3D object reconstruction task, as the DETR-style head fails to pick up any training signal if the network is trained without it.

\mypar{2D foreground-background (FG/BG) segmentation}
As a second auxiliary task, we predict a 2D \emph{amodal} segmentation mask in each view, for all objects together.
A pixel belongs to the mask if it lies on any object in the view, regardless of occlusion.
We use transposed convolutions, combined with non-linearities and normalization layers, to up-sample the pixel stream output $P_i^n$ of the last 2D $\Leftrightarrow$ 3D block to the original input resolution (16-fold).
We supervise using the binary cross entropy loss.
We create the amodal masks by rasterizing the combined geometry of all ground-truth 3D objects into each view.
Predicting amodal masks enhances the backbone's 3D understanding of the world.
In general, the amodal mask is ill-defined for occluded regions in a single image. It becomes well defined with multiple views however, if some of them observe the object behind the occluder. Hence, this FG/BG task pushes our network to reason about geometric relations across multiple views.

\mypar{Novel View Synthesis}
While we focus on multi-object 3D reconstruction, our backbone and the scene-level representation it outputs can be used for other 3D tasks as well. In the appendix, we provide qualitative results for Novel View Synthesis, which builds upon \OURS{}'s backbone.

%% file: 04_experiments.tex
\vspace{-2mm}
\section{Experiments}
\label{sec:experiments}
\vspace{-2mm}

\begin{table*}[t]
\small
\centering
\begin{minipage}{\linewidth}
\resizebox{\columnwidth}{!}{
\small
\begin{tabular}{c|l||>{\columncolor[gray]{0.9}}c>{\columncolor[gray]{0.9}}c|ccccccccccc}
\textbf{Family}  &\textbf{Method} &
\rott{\textbf{class avg.}} & \rott{\textbf{global avg.}} & \rott{bathtub}& \rott{bookshelf}& \rott{cabinet}& \rott{chair}& \rott{display}& \rott{sofa}& \rott{table}& \rott{trashbin}& \rott{other} \\
\hline
\multirow{2}{*}{\begin{tabular}[c]{@{}c@{}}Single-frame\\baselines\end{tabular}}
    & Mask2CAD~\cite{kuo20eccv} +avg   & 2.5  & 3.5  & 0.0  & 1.9  & 1.5  & 6.8  & 3.7  & 2.7  & 1.4  & 3.0  & 1.2 \\
    & Mask2CAD~\cite{kuo20eccv} +pred   & 11.6 & 16.0 & 8.3  & 3.8  & 5.4  & 30.9 & 17.3 & 5.3  & 7.1  & 25.9 & 0.5 \\
\hline
\multirow{3}{*}{\begin{tabular}[c]{@{}c@{}}Multi-Frame\\Methods\end{tabular}}
    & MVS~\cite{duzceker21deepvideomvs} + RGB-D fitter~\cite{avetisyan19iccv}  & 18.8 & 21.7 & 15.8 & 8.5 & 17.3 & 34.3 & 25.7 & 15.0 & 10.9 & 35.8 & 6.1 \\
    & ODAM~\cite{li21odam}             & 25.6 & 29.2 & 24.2 & 12.3 &13.1. & 42.8 & 36.6 & 28.3 & 31.1 & 42.2 & 0.0 \\
    & Vid2CAD~\cite{maninis20vid2cad}      & 30.7 & 38.6 & 28.3 & 12.3 & 23.8 & 64.6 & 37.7 & 26.5 & 28.9 & 47.8 & 6.6 \\
    & \OURS                                & \textbf{36.2} & \textbf{43.0} & 19.2 & 34.4 & 36.2 & 59.3 & 30.4  & 44.2 & 42.5 & 31.5 & 27.8 \\\hline
\end{tabular}%
}
\end{minipage}
\vspace{0mm}
\caption{ \small
Quantitative results on the Scan2CAD~\cite{avetisyan19cvpr} dataset using the original Scan2CAD metrics.
Results for Mask2CAD variants, MVS+RGB-D fitter, and Vid2CAD are as reported in~\cite{maninis20vid2cad}.
ODAM originally reports in another metric (Tab.~\ref{tbl:odam_metric}).
We re-evaluate in the Scan2CAD metrics based on model outputs provided to us by the authors.
Note that ODAM was not trained to predict the `other` class. When excluding it from the metrics, ODAM achieves class avg. of 28.8\% and global avg. of 33.5\%.
}
\label{tbl:exp}
\vspace{-6mm}
\end{table*}

\myfirstpar{Datasets and evaluation metrics}
We evaluate our method on Scan2CAD~\cite{avetisyan19cvpr}, following their protocol and evaluation metrics.
Concretely, we use videos from ScanNet~\cite{dai17cvpr}, 3D CAD models from
ShapeNetCore~\cite{chang15arxiv}, and annotations that connect the two from
Scan2CAD~\cite{avetisyan19cvpr}. ScanNet provides videos of rich indoor scenes
with multiple objects in complex spatial arrangements.
ShapeNetCore provides CAD models from 55 object classes, in a canonical
orientation within a class.
Scan2CAD provides manual 9-DoF alignments of
ShapeNetCore models onto ScanNet scenes for 9 super-classes.

We use these datasets both for training and evaluation.
During training, we consider all ScanNet videos in the official train split whose scenes have
Scan2CAD annotations (1194 videos).
We evaluate on the 306 videos of the validation set, containing a total of 3184 aligned 3D objects.
We quantify performance using the original Scan2CAD metrics~\cite{avetisyan19cvpr} and the metrics introduced in ODAM~\cite{li21odam}.
In the Scan2CAD metrics, a ground-truth 3D object is considered accurately detected if one of the objects output by the model matches its class and pose alignment (passing three error thresholds at the same time: $20$\%
scale, $20^\circ$ rotation, $20$cm translation).
We report accuracy averaged over classes (‘class avg.’) as well as over all object instances (‘global avg’).
In the metrics of~\cite{li21odam}, an object is considered accurately detected 
if the Intersection-over-Union (IoU) of its oriented 3D bounding box to a ground-truth box of the same class is above a predefined threshold.
We report precision, recall, and F1 score.
Finally, the dataset also provides dense 3D meshes for the scene produced using a dedicated depth sensor. We ignore this data, both at training and test time (in contrast to some previous works which rely on it~\cite{avetisyan19cvpr,avetisyan19iccv,avetisyan20eccv,izadinia20cvpr,shan2021ellipsdf}).

\renewcommand{\thempfootnote}{\arabic{mpfootnote}}
\begin{table}[t]
    \centering
    \begin{minipage}{\linewidth}
    \resizebox{\columnwidth}{!}{
    \begin{tabular}{l|c@{\hskip 4mm}c@{\hskip 2mm}c@{\hskip 2mm}c@{\hskip 2mm}c|}
       Prec./Rec./F1 & MOLTR~\cite{li2020arxiv} & ODAM~\cite{li21odam} & Vid2CAD~\cite{maninis20vid2cad} & ImVoxelNet~\cite{rukhovich2021imvoxelnet} & \OURS{} \\
        \hline
        @IoU$>0.25$ & 54.2/55.8/55.0 & 64.7/58.6/61.5 & 56.9/55.7/56.3 & 52.9/53.2/53.0 & \textbf{65.4}/\textbf{61.8}/\textbf{63.6} \\ 
        @IoU$>0.5$  & 15.2/17.1/16.0 & 31.2/28.3/29.7 & 34.2/33.5/33.9 & 17.0/17.1/17.0 & \textbf{41.9}/\textbf{39.6}/\textbf{40.7} \\
    \end{tabular}
    }
    \end{minipage}
    \vspace{1mm}
    \caption{ \small
    Quantitative results on Scan2CAD using the ODAM~\cite{li21odam} metrics.
    To evaluate \OURS{}, we derive an oriented 3D box by using the 3D transformations predicted by the model.
    \OURS{} outperforms all other works, especially at the stricter IoU threshold (@IoU$>0.5$), showing it produces particularly accurate object poses.
    Note that for Vid2CAD we report the updated results from \url{https://github.com/likojack/ODAM} (which match exactly the Vid2CAD paper \cite{maninis20vid2cad}).
    Also note that ImVoxelNet outputs axis-aligned boxes, which hinders its performance at high IoU thresholds.
    Finally, results for MOLTR and ODAM are as reported in~\cite{li21odam}.
    }
    \vspace{-6mm}
    \label{tbl:odam_metric}
\end{table}

\mypar{Training details}
We implement our model in PyTorch~\cite{paszke2019pytorch}.
We train on 20 frames per video, using 16-bit float arithmetic. This allows us to fit one video on a GPU with 16GB of memory. We use 8 GPUs in total, resulting in a batch size of 8.
We train \OURS{} in three stages. We first train just the backbone for 224k steps (1500 epochs) on the task of predicting 3D occupancy (Sec.~\ref{sec:heads}). We then enable all other tasks except the shape predictor and we train for another 239k steps (1604 epochs). Finally, we train just the shape predictor for another 5k steps (17 epochs), after freezing the rest of the network parameters.
We use the AdamW~\cite{loshchilov2017decoupled} optimizer, with a learning rate of $10^{-4}$ and weight decay $5\cdot10^{-2}$.

\mypar{Compared methods}
We compare \OURS{} against Vid2CAD~\cite{maninis20vid2cad}, ODAM~\cite{li21odam},
MOLTR~\cite{li2020arxiv}, and ImVoxelNet~\cite{rukhovich2021imvoxelnet}, four 
recent methods for 3D object pose estimation and detection from RGB videos.

We further compare to two baselines that process frames individually, defined
by~\cite{maninis20vid2cad}. These extend Mask2CAD~\cite{kuo20eccv}, which in its
original form does not predict the 3D depth nor the scale of the object. The
first baseline, `Mask2CAD +avg', estimates an object's depth and scale by taking
the average over its class instances in the training set. The second baseline,
`Mask2CAD +pred', predicts the scale of the actual object in the image (and then
derives its depth from it).
Both baselines aggregate 3D object predictions across all video frames and remove duplicates that occupy the same volume in 3D.

Several previous methods report strong results on Scan2CAD by using
a dedicated RGB-D depth sensor to acquire a dense 3D point-cloud of the scene.
Those methods have an intrinsic advantage and operate by directly fitting CAD models on
the scene's 3D point cloud~\cite{avetisyan19cvpr,avetisyan19iccv,avetisyan20eccv,izadinia20cvpr}.
Instead, our method only uses the RGB frames.
Hence, we compare to a strong alternative method, defined
in~\cite{maninis20vid2cad}, that replaces the input of the best RGB-D fitting
method~\cite{avetisyan19iccv} with 3D point-clouds generated by the
state-of-the-art multi-view stereo method DVMVS~\cite{duzceker21deepvideomvs}.
We train DVMVS on ScanNet, and re-train~\cite{avetisyan19iccv} on its output.

\begin{figure}[t]
	\centering
		\vspace{-2mm}
	\includegraphics[width=1\linewidth]{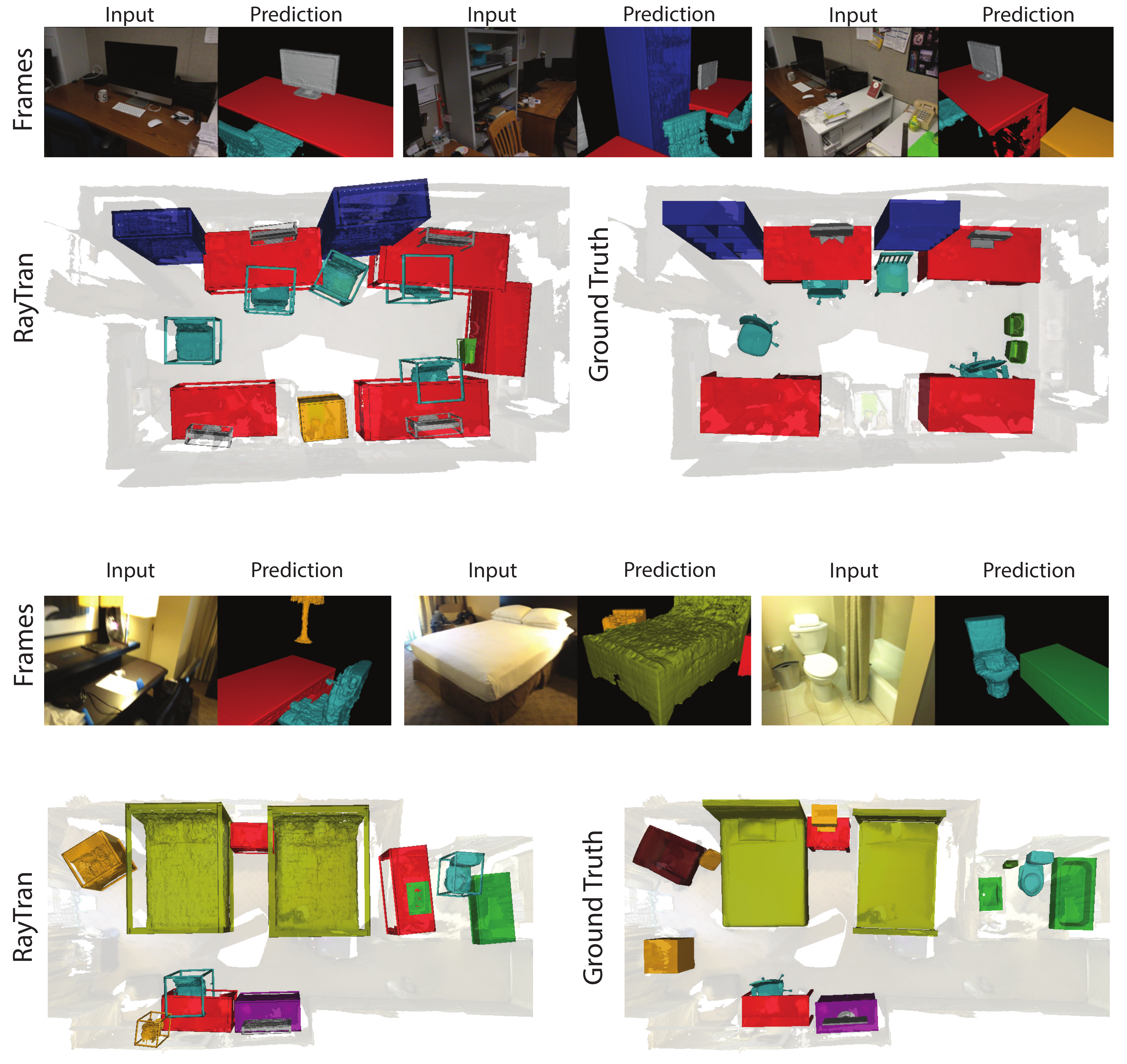} \\[1mm]
		\vspace{-2mm}
	\caption{\small
		\textbf{Qualitative Results (top-view, with frame overlays)}:
    	We show the 3D pose estimation (as oriented boxes) and shape reconstruction outputs of \OURS{} against the ground truth, from the top and from the viewpoint of the images.
		The objects are colored by class. 
		We are able to reconstruct complex scenes in a single pass.
	}
	\label{fig:qualitative_with_frames}
\end{figure}

\begin{figure}[t]
	\centering
	\includegraphics[width=1\linewidth]{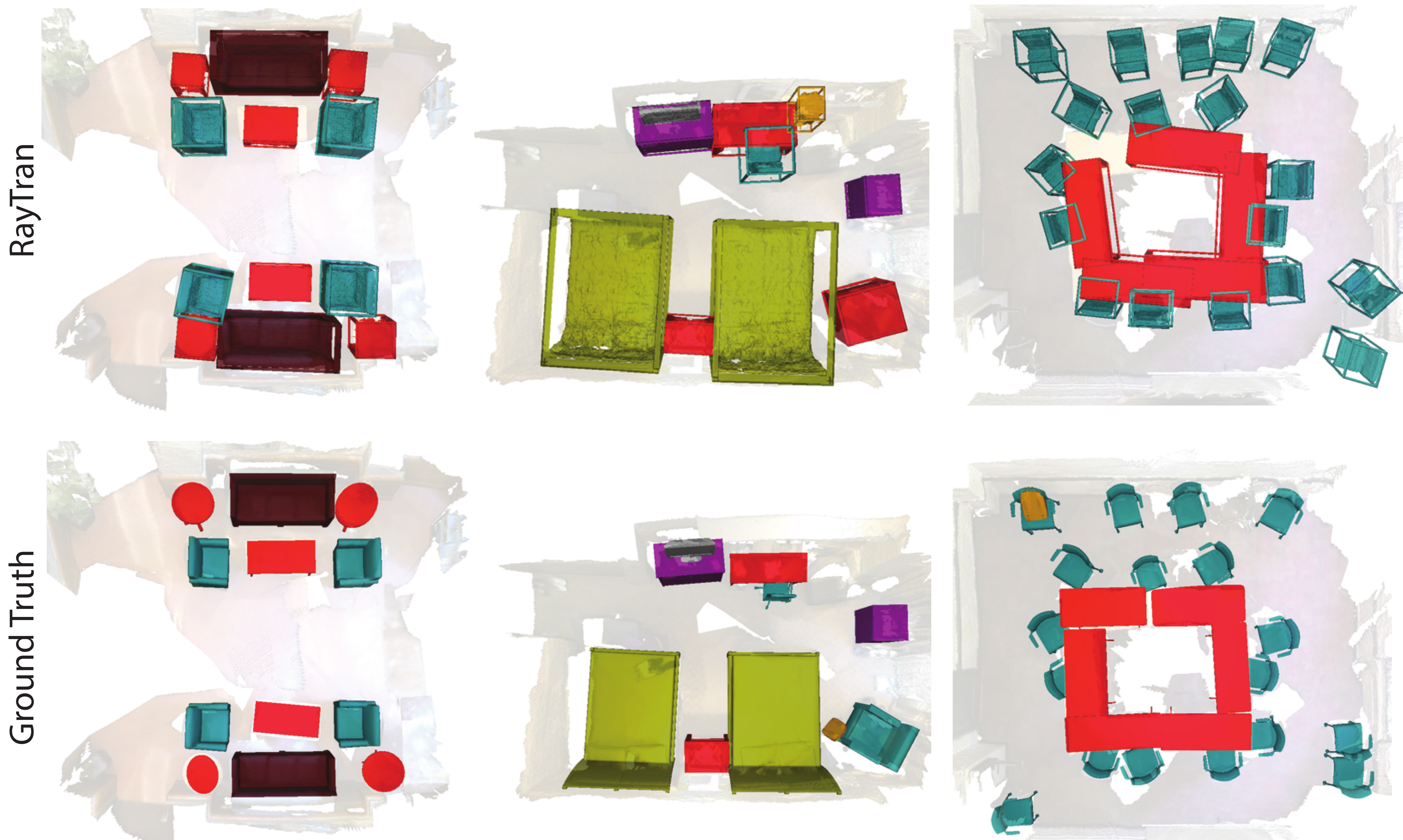} \\[1mm]
	\caption{\small
	\textbf{Additional qualitative Results (top-view)}:
		We show the 3D pose estimation and shape reconstruction outputs for 3 additional scenes. For 
		each detected object we visualize its 3D oriented bounding box, as well as its reconstructed 
		mesh.
	}
	\label{fig:qualitative_top_view}
\end{figure}

\mypar{Main results}
Tab.~\ref{tbl:exp} shows the results in the Scan2CAD metrics.
\OURS{} outperforms both single-frame baselines as well as the `MVS + RGBD
fitter' combination by a wide margin ($+33.7\%$, $+24.6\%$, $+17.4\%$ class avg.
accuracy respectively).
\OURS{} also outperforms both competitors that align CAD model to RGB videos
but rely on tracking:
Vid2CAD~\cite{maninis20vid2cad} ($+5.5\%$) and ODAM~\cite{li21odam} ($+10.6\%$).
Importantly, \OURS{} is also much simper in design, as \cite{maninis20vid2cad,li21odam} consist of multiple disjoint steps (object detection, tracking, multi-view optimization).
Fig.~\ref{fig:qualitative_with_frames} and Fig.~\ref{fig:qualitative_top_view} illustrate qualitative 
results for our method.

Looking at individual categories, we obtain the best result on 5 out of 9, and in
particular on the "other" category, which is hard for methods based on
retrieving CAD models~\cite{maninis20vid2cad,avetisyan19cvpr,avetisyan19iccv}.
Our method instead predicts 3D shapes as voxel grids which helps to
generalize better and to adapt to the large variety of object shapes in this
catch-all category.
On `trashbin' we do moderately worse, possibly because of the relatively coarse voxel resolution of the backbone representation.

For completeness, we also compare to methods \cite{avetisyan19cvpr,avetisyan19iccv} in their original form, i.e. fitting CAD models to high-quality dense RGB-D scans.
Surprisingly, \OURS{} (36.2\%/43.0\%) improves over~\cite{avetisyan19cvpr} (35.6\%/31.7\%),
despite using only RGB video as input.
While the state-of-the-art~\cite{avetisyan19iccv} performs even better (44.6\%/50.7\%), this family
of methods are limited to videos acquired by RGB-D sensors.

\begin{figure}[t]
\centering
\vspace{-4mm}
\includegraphics[width=1\linewidth]{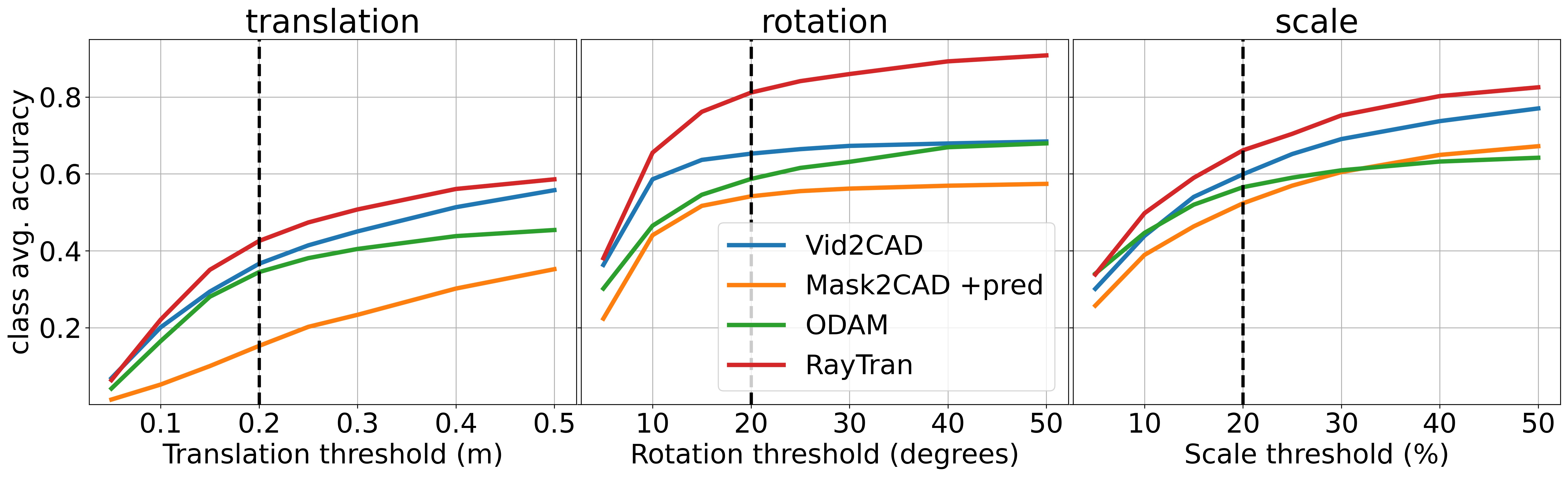} \\[1mm]
\vspace{-2mm}
\caption{ \small
  \textbf{Transformation type ablation}: Class-avg accuracy as a function of the evaluation threshold (the vertical dotted line shows the default value, used in Tab.~\ref{tbl:exp}). We examine each transformation type separately. \OURS{} achieves better accuracy than Vid2CAD, ODAM, and `Mask2CAD +pred' on all transformation types.
}
\label{fig:scan2cad_metric_separate}
\vspace{-4mm}
\end{figure}

Fig.~\ref{fig:scan2cad_metric_separate} reports accuracy for each
transformation type separately (translation, rotation, and scales).
Our method predicts all transformation types better than Vid2CAD, ODAM, and the best single-frame baseline Mask2CAD+pred.
As objects are considered accurately detected only when
passing all 3 thresholds \emph{simultaneously}, improving translation is the
biggest avenue for improving our overall quantitative results (Tab.~\ref{tbl:exp}).

Tab.~\ref{tbl:odam_metric} reports results in the ODAM metrics, which allow us to compare to MOLTR~\cite{li2020arxiv} and ImVoxelNet~\cite{rukhovich2021imvoxelnet}.
We choose an object score threshold to maximize the F1 score on the val set for methods that predict object scores (\OURS{}, Vid2CAD, ImVoxelNet), following the practice of~\cite{li21odam}.
\OURS{} outperforms all four methods~\cite{li21odam,li2020arxiv,rukhovich2021imvoxelnet,maninis20vid2cad} at both IoU thresholds.
ImVoxelNet~\cite{rukhovich2021imvoxelnet} reports results on ScanNet, not on Scan2CAD. To compare properly we use their publicly available source code and re-train on Scan2CAD. The original code only outputs axis-aligned object boxes on ScanNet (and hence on Scan2CAD, which is derived from it). This prevents comparison on the Scan2CAD metrics, as we cannot compute precise rotation and scale components. Finally, predicting box rotation could potentially improve the results in the ODAM metrics.

\begin{table}[t]
\centering
\resizebox{\columnwidth}{!}{
\begin{tabular}{l@{\hskip 2
mm}|c@{\hskip 5mm}c|c@{\hskip 5mm}c}
Method.                            & extra input & auxiliary tasks                 & class avg. & global avg. \\\hline
\OURS{}                            & -           & 3D occupancy + 2D FG/BG seg.    & 36.2       & 43.0 \\
\OURS{} w/o FG/BG.                 & -           & 3D occupancy                    & 33.8       & 40.1 \\ \hline
\OURS{} + GT masks                 & 2D GT masks & 3D occupancy + 2D FG/BG seg.    & 47.6       & 52.5 \\
\end{tabular}
}
\vspace{2mm}
\caption{ \small
Effects of object segmentation in~\OURS.
Without FG/BG segmentation as an auxiliary task, the network performs worse (first two rows).
If we grant the model the ground-truth segmentation masks as input, results substantially improve, highlighting how future progress on automatic 2D segmentation will benefit our work too (last row).
In all cases, the model is trained with the main 3D object pose/shape estimation loss (Sec. \ref{sec:heads}), in addition to the auxiliary losses listed here.}
\label{tbl:ablation}
\vspace{-6mm}
\end{table}

\mypar{Ablation: 2D FG/BG segmentation as auxiliary task}
Our model predicts amodal masks, which reinforces the backbone's 3D understanding. Pixels where the object is occluded can only be predicted correctly in 2D as part of the amodal mask by relying on signal from other frames, via the global 3D representation.
To support this claim, we trained a version of our model where we disabled the
2D FG/BG segmentation auxiliary task of Sec.~\ref{sec:heads}.
This reduces class-avg accuracy by -2.4\% (36.2\% vs. 33.8\, first two rows of Tab.~\ref{tbl:ablation}).

\mypar{Ablation: Perfect segmentation}
Our model performs both 2D and 3D analysis. The main challenge in the 2D analysis is pixelwise segmentation in the input frames. We explore here what would happen if our model were granted perfect object segmentation as input.
We train a model which inputs a binary mask as a 4th channel, in addition to RGB.
A pixel in the mask is on if it belongs to any object of the 9 classes annotated in Scan2CAD, and 0 otherwise.
This augmented model improves class-avg accuracy by $+11.4\%$ (reaching $47.6\%$), and
global-avg by $+9.5\%$ (reaching $52.5\%$, Tab.~\ref{tbl:ablation} last row).
Hence, as research on 2D segmentation improves, so will our model's 3D scene understanding ability.

\mypar{Ablation: Number of objects in the scene}
By design, our network cannot predict more object instances than the query slots
in the DETR head (64). Moreover, typically only about $30\%$ of all query slots
bind to an actual object~\cite{carion2020end} in scenes containing many objects.
This limits recall on such scenes, which sometimes do occur in Scan2CAD.
Carion et al~\cite{carion2020end} believe that query slots tend to bind to
fixed spatial regions, regardless of the content of a test image, causing this limitation.
We operate in 3D, which likely exacerbates it because we need many more queries to cover
the 3D space.

To understand the effect of this phenomenon on our model's performance, we
evaluate here on 3 subsets of Scan2CAD's val split, containing scenes with \emph{at most} 10, 20, and 30
objects respectively. This reduces the number of objects undetected by the fixed
64 query slots in our DETR head.

The class-avg accuracy of \OURS{} indeed improves in scenes containing fewer objects (from $36.2\%$ in all scenes, up to $37.4\%$ in scenes with $<10$ objects).
The accuracy of the best previous method Vid2CAD instead remains constant.
In scenes with at most 10 objects, we outperform Vid2CAD by $6.7\%$ ($37.4\%$ vs. $30.7\%$),
which is a larger difference than on all scenes ($5.5\%$: $36.2\%$ vs. $30.7\%$).
Hence, DETR's limitation is affecting our model as well and improving upon it will improve our overall performance.

\mypar{Ablation: Number of input frames}
Our model can process a variable number of input frames per video. We use 20 frames at training time to limit memory requirements. In all experiments so far, we used 96 frames at inference time, as using more frames improves coverage of the 3D volume of the scene and hence accuracy of the output.
To support this claim, we now reduce the number of frames at inference time. With 48 frames class-avg. falls by -0.4\%. Worse yet, if inference were constrained to 20 frames as during training, then performance would drop by -3.3\%.
This highlights the value of our model's ability to input a variable number of frames.

%% file: 05_conclusions.tex
\section{Conclusions}
\label{sec:conclusions}

We presented \OURS{}, a novel backbone architecture for 3D scene reconstruction from RGB video frames, that uses transformers for unprojecting 2D features and consolidating them into a global 3D representation. We introduced the ray-traced sparse transformer block, which enables feature sharing between the 2D and 3D network streams, in a computationally feasible way on current hardware.
We use this architecture to perform 3D object reconstruction for the full scene by combining it with a DETR-style network head. Our architecture can reconstruct the whole scene in a single pass, is end-to-end trainable, and does not rely on tracking.
We perform experiments on the Scan2CAD benchmark,
where \OURS{} outperforms 
(1) recent state-of-the-art methods \cite{maninis20vid2cad,li21odam,li2020arxiv,rukhovich2021imvoxelnet} for 3D object pose estimation from RGB videos; and
(2) a strong alternative method combining Multi-view Stereo \cite{duzceker21deepvideomvs} with RGB-D CAD alignment \cite{avetisyan19iccv}.

%% file: 06_appendix.tex
\section*{Appendix A: Preliminary results on novel view synthesis}
In the previous sections, we showed that the \OURS{} backbone is an effective approach for 3D pose estimation and shape reconstruction from videos, and leads to state-of-the-art results. On top of the main task, the architecture enables two additional auxiliary tasks: 3D occupancy prediction, and 2D foreground-background segmentation.

Similarly, the features created by \OURS{} and their dual $3D \Leftrightarrow 2D$ interpretation enable other different tasks that use posed images as input and require geometric reasoning.
In this section we re-purpose the backbone for the task of novel view synthesis.
Despite being a fundamentally different task, we are able to approach novel view synthesis with minimal modifications to the same end-to-end trainable architecture.

We enable novel view synthesis by projecting the 3D voxel grid $V$ to a $30 \times 40$ 2D grid $P_i$ aligned with the novel (query) view, by using a new camera pose. To this end, we use an additional $3D \rightarrow 2D$ transformer block.
We then use a simple decoder comprising of transpose convolutions with non-linearities and normalization layers to recover images at the original input resolution ($480 \times 640$), and we supervise with the MSE loss.
We ask our model to predict new query views, unseen during training. 
The encoder-decoder setup is such that the encoder does not have access to the query frame parameters, forcing it to infuse the 3D representation with information relevant to all possible query viewpoints.

\begin{figure}[h]
	\centering
	\includegraphics[width=1\linewidth]{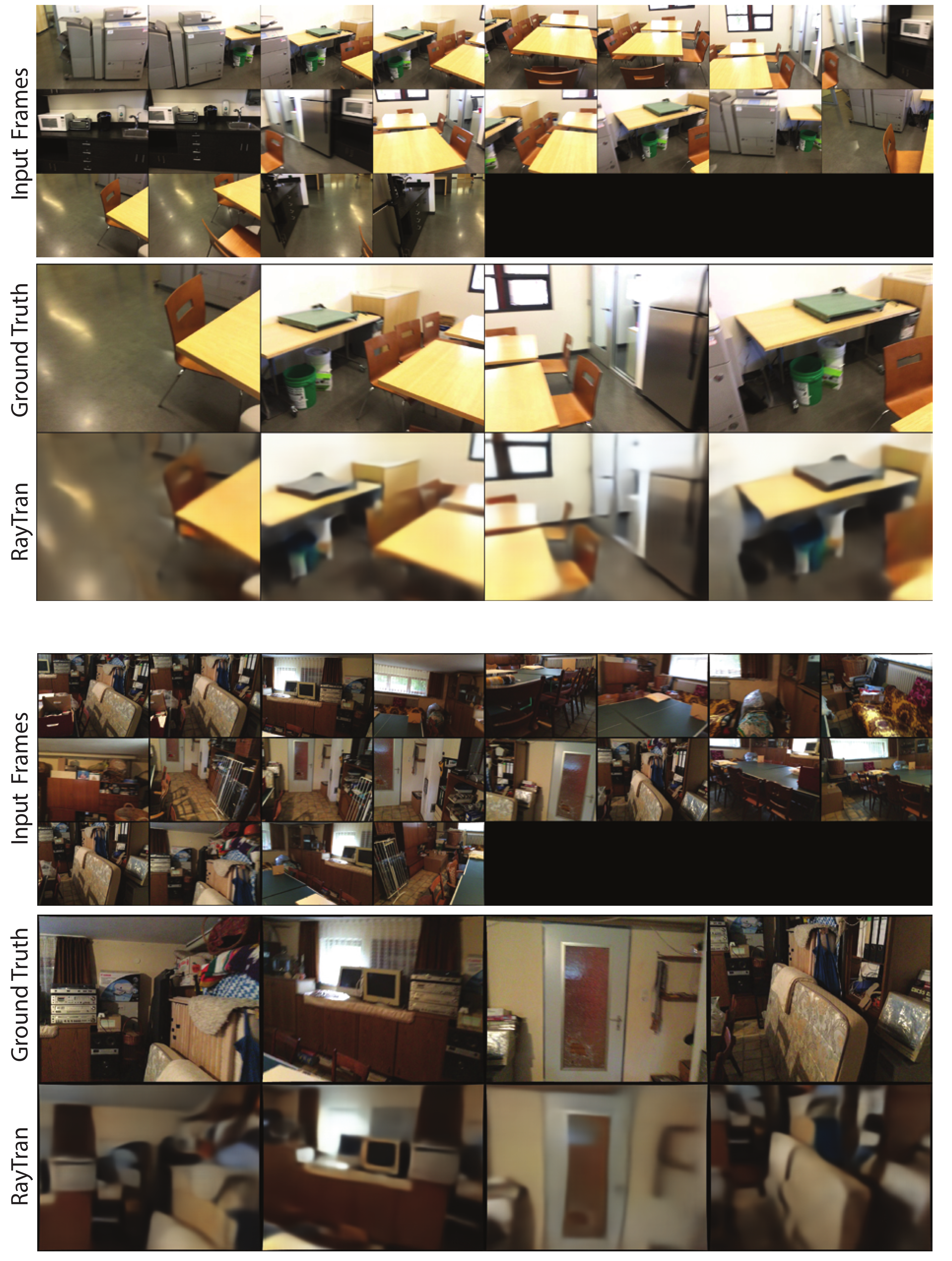} \\[1mm]
	\vspace{-4mm}
	\caption{\textbf{Novel view synthesis (NVS).}
		Qualitative results for 2 different scenes from the test set.
		For each scene, we show the input frames to the network (top), the ground-truth views from new query camera viewpoints (middle), and the corresponding views predicted by the novel view synthesis head on top of the \OURS{} backbone (bottom).
		The $2D \Rightarrow 3D$ attention in our model recovers the view's features
		at a resolution of $30 \times 40$, which is then upsampled to $480 \times 640$.
		}
	\label{fig:nvs_1}
\end{figure}

\begin{figure}[h]
	\centering
	\includegraphics[width=1\linewidth]{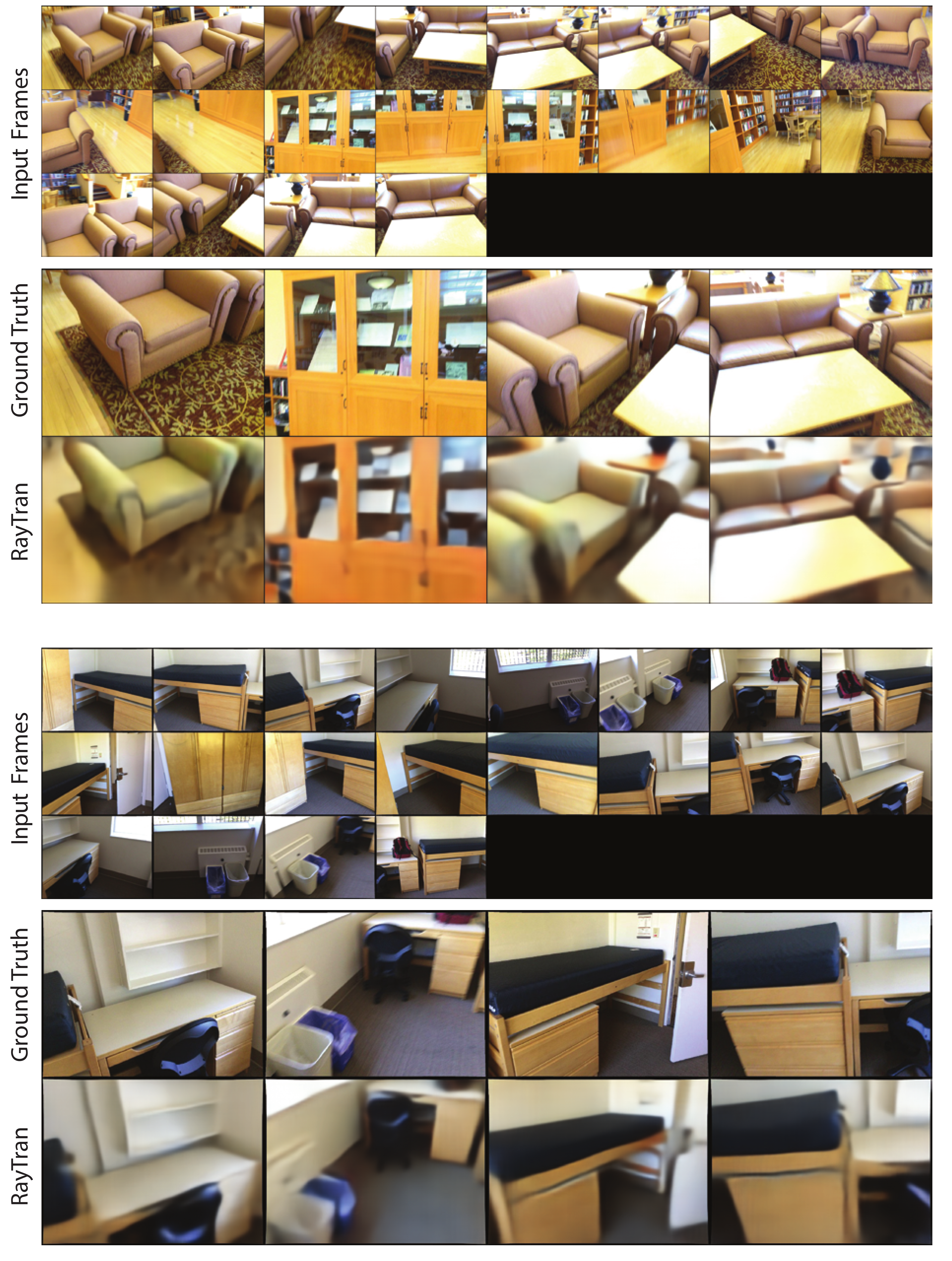} \\[1mm]
	\vspace{-4mm}
	\caption{\textbf{More results on novel view synthesis (NVS).}
		Qualitative results for 2 different scenes from the test set.
		For each scene, we show the input frames to the network (top), the ground-truth views from new query camera viewpoints (middle), and the corresponding views predicted by the novel view synthesis head on top of the \OURS{} backbone (bottom).
		The $2D \Rightarrow 3D$ attention in our model recovers the view's features
		at a resolution of $30 \times 40$, which is then upsampled to $480 \times 640$.
		}
	\label{fig:nvs_2}
\end{figure}

We show preliminary qualitative results in Fig.~\ref{fig:nvs_1} and Fig.~\ref{fig:nvs_2}.  
Our method predicts the overall structure of the images rather well, which suggests that it builds features for sufficient geometric reasoning.
It is less accurate in terms of high frequency details, likely because the resolution of the 2D feature grid is too low for the simple network that we use for upscaling.
In contrast to NeRF-like architectures~\cite{mildenhall2020nerf}, however, it does not require any training at test time.

\section*{Appendix B: Additional Qualitative Results on 3D pose and shape.}
\label{sec:additional_qualitative}
\begin{figure}[h]
	\centering
	\vspace{-2mm}
	\includegraphics[width=1\linewidth]{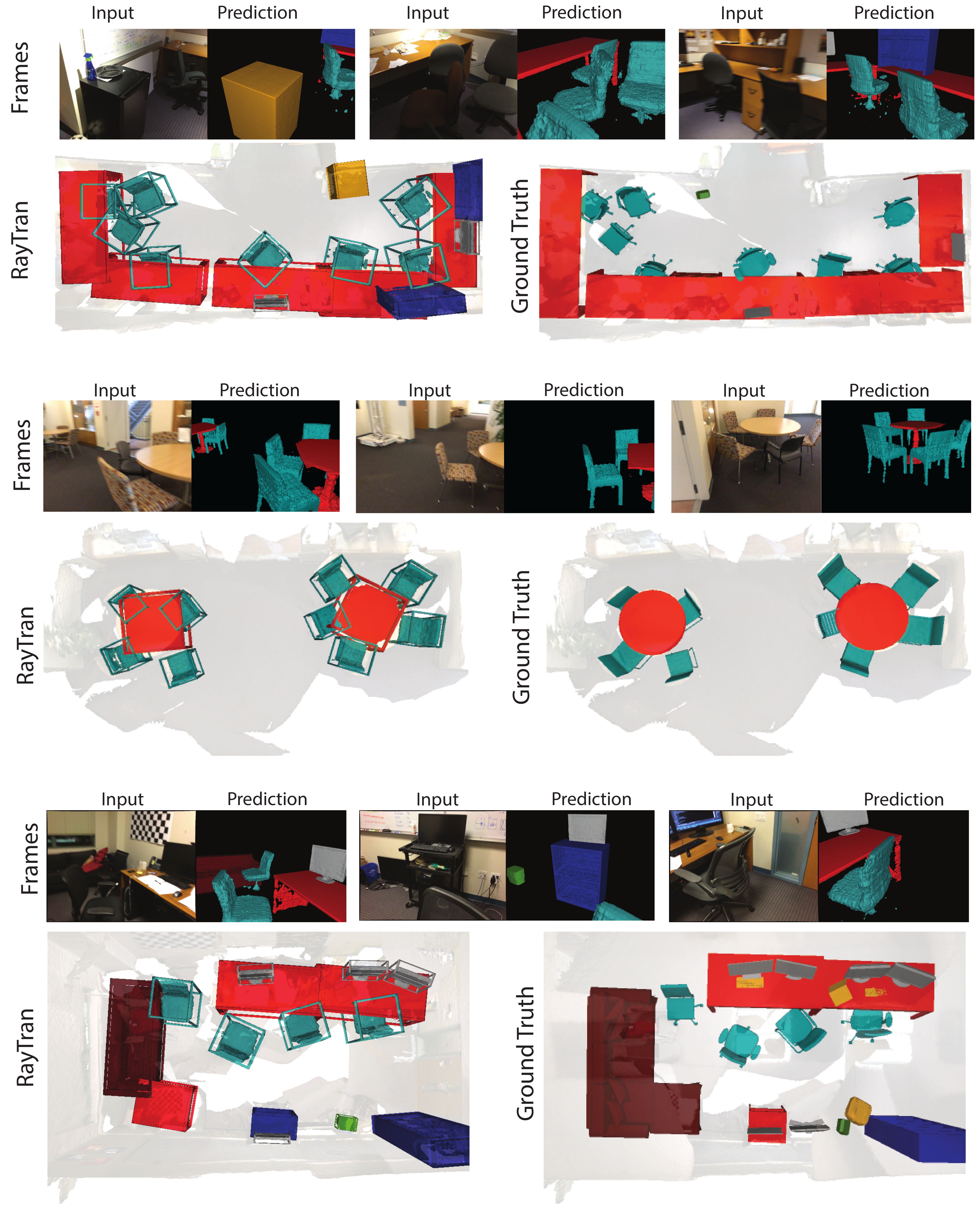} \\[1mm]
	\vspace{-0mm}
	\caption{\small
		\textbf{Qualitative Results for 3D pose and shape (with frame overlays)}:
		We show 3D pose estimation and shape reconstruction outputs of \OURS{} against the ground truth. 
		We also show the results from the viewpoint of the images.
		The objects are colored by class.
		Notice that \OURS{} predicts objects that are omitted from the ground truth but appear in the video (eg. cabinet and the bookshelf of top row).
	}
	\label{fig:supp_qualitative_with_frames}
\end{figure}

\begin{figure}[h]
	\centering
	\vspace{-2mm}
	\includegraphics[width=1\linewidth]{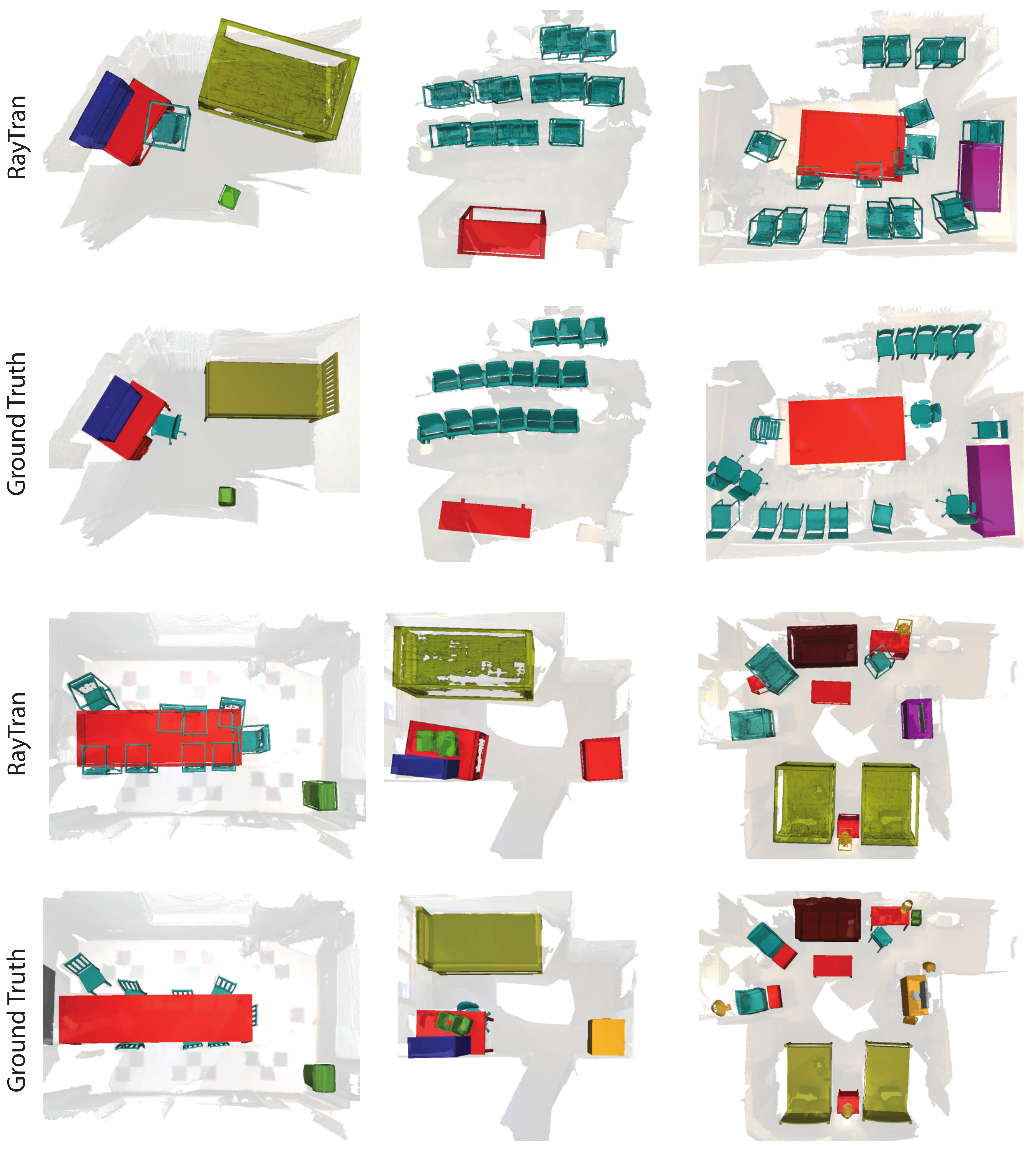} \\[1mm]
	\vspace{-0mm}
	\caption{\small
		\textbf{More qualitative Results for 3D pose and shape (top-view)}:
		We show the 3D pose estimation and shape reconstruction outputs for 6 example scenes. For each detected object we visualize its 3D oriented bounding box, as well as its reconstructed mesh.
		For reference, we also show the ground-truth alignments, as well as the \mbox{RGB-D} scan of the scene in the background (which we do not use).
		\OURS{} is able to reconstruct objects of very complicated scenes with densely placed chairs (top row, columns 2 and 3), and scenes with many different classes. It also reconstructs large objects that are often truncated in every individual frame of the video due to their size, and thus need multiple frames to be reconstructed (tables of top row column 3, bottom row column 1).
        Failure cases include predicting an object from a different class (eg. table instead of a similar-looking cabinet of bottom row, column 2), or missing objects entirely (eg. the left lamp of bottom row, column 3).
	}
	\label{fig:supp_qualitative_top_view}
\end{figure}

Fig.~\ref{fig:supp_qualitative_with_frames} and Fig.~\ref{fig:supp_qualitative_top_view} illustrate additional results of \OURS{} in Scan2CAD for the task of 3D pose estimation and shape reconstruction, including successful reconstructions, as well as typical failure cases.

